\documentclass[twocolumn]{cinc}

\usepackage{graphicx}
\usepackage{pgf,tikz}
\usepackage{pgfplots}

\usepackage{amsmath}
\usepackage{url}

\usepackage{amsfonts}
\usepackage{amssymb}
\usepackage{amsbsy}
\usepackage{dsfont}

\usepackage{times}
\usepackage{default}
\usepackage{color}

\usepackage{tikz}
\usetikzlibrary{shapes.geometric}
\usetikzlibrary{shapes.arrows}
\usepackage{array}

\usepackage{subcaption}
\usepackage{multirow}

\newcommand{\bpar}[1]{\textbf{#1}}

\begin{document}
\bibliographystyle{cinc}

% Keep the title short enough to fit on a single line if possible.
% Don't end it with a full stop (period).  Don't use ALL CAPS.
\title{Convolutional Recurrent Neural Networks for Electrocardiogram Classification}

% Both authors and affiliations go in the \author{ ... } block.
% List initials and surnames of authors, no full stops (periods),
%  titles, or degrees.
% Don't use ALL CAPS, and don't use ``and'' before the name of the
%  last author.
% Leave an empty line between authors and affiliations.
% List affiliations, city, [state or province,] country only
%  (no street addresses or postcodes).
% If there are multiple affiliations, use superscript numerals to associate
%  each author with his or her affiliations, as in the example below.

\author {Martin Zihlmann, Dmytro Perekrestenko, Michael Tschannen \\
\ \\ 
Dept. IT \& EE, ETH Zurich, Switzerland
}

\maketitle

% LaTeX inserts the ``Abstract'' heading in the proper style and
% sets the text of the abstract in italics as required.
\begin{abstract}

We propose two deep neural network architectures for classification of arbitrary-length electrocardiogram (ECG) recordings and evaluate them on the atrial fibrillation (AF) classification data set provided by the PhysioNet/CinC Challenge 2017. The first architecture is a deep convolutional neural network (CNN) with averaging-based feature aggregation across time. The second architecture combines convolutional layers for feature extraction with long-short term memory (LSTM) layers for temporal  aggregation of features. As a key ingredient of our training procedure we introduce a simple data augmentation scheme for ECG data and demonstrate its effectiveness in the AF classification task at hand. The second architecture was found to outperform the first one, obtaining an $F_1$ score of $82.1$\% on the hidden challenge testing set.

%Center the word Abstract and print it in Times New Roman 11-point size and in bold.
%
%% Of course, you must insert blank lines
%% between the paragraphs of your LaTeX input file, since this is the
%% only way to indicate paragraph boundaries.  Make sure that no blank
%% lines appear between paragraphs in the formatted output, however.)
%% 
%% 
%The abstract and all other text (except where indicated in the instructions) is printed in a 10-point size. The text of the abstract (and only the abstract) should be in italics. Begin all paragraphs with a 1 pica (4 mm, 1/6 inch) indentation.
%
%When you come to additional paragraphs, do not add blank line spaces between paragraphs, here or anywhere else in your article.
%
%The abstract with its heading should not be more than 100 mm long. This is equivalent to 25 lines of text. Leave 2 line spaces at the bottom of the abstract before continuing with the next heading.

\end{abstract}
% LaTeX inserts the extra space here automatically.
\vspace{-2.5mm}

\section{Introduction}

We consider the task of atrial fibrillation (AF) classification from single lead electrocardiogram (ECG) recordings, as proposed by the PhysioNet/CinC Challenge 2017 \cite{clifford2017af}. AF occurs in 1-2\% of the population, with incidence increasing with age, and is associated with significant mortality and morbidity \cite{lip2016atrial}. Unfortunately, existing AF classification methods fail to unlock the potential of automated AF classification as they suffer from poor generalization capabilities incurred by training and/or evaluation on small and/or carefully selected data sets.
 
In this paper, we propose two deep neural network architectures for classification of arbitrary-length  ECG recordings and evaluate them on the AF classification data set provided by the PhysioNet/CinC Challenge 2017. The first architecture is a 24-layer convolutional neural network (CNN) with averaging-based feature aggregation across time. The second architecture is a convolutional recurrent neural network (CRNN) that combines a 24-layer CNN with a 3-layer long-short term memory (LSTM) network for temporal  aggregation of features. CNNs have the ability to extract features invariant to local spectral and spatial/temporal variations, and have led to many breakthrough results, most prominently in computer vision \cite[Chap. 9]{goodfellow2016deep}. LSTM networks, on the other hand, were shown to effectively capture long term temporal dependencies in time series \cite[Chap. 10]{goodfellow2016deep}. As a key ingredient of our training procedure we introduce a simple yet effective data augmentation scheme for the ECG data at hand.
 
\bpar{Related work:} Our network architectures are loosely inspired by \cite{lipton2015learning,bashivan2015learning,cakir2017convolutional}. More specifically, a CRNN for polyphonic sound detection was proposed in \cite{cakir2017convolutional}. Here, unlike in AF classification where one has to infer a single label per ECG, the input audio sequence is mapped to sequences labels, inferring the sound events as a function of time. Work \cite{bashivan2015learning} employs a CRNN for mental state classification from electroencephalogram (EEG) data. In \cite{lipton2015learning}, LSTM networks are used for multilabel classification of diagnoses in electronic health recordings. Shortly before finalizing this work, we became aware of the preprint \cite{rajpurkar2017cardiologist}, which proposes a deep CNN architecture for arrhythmia detection in ECGs, but unlike in the classification problem considered here, maps the ECG signal to a sequence of rhythm classes. Finally, we refer to \cite{larburu2011comparative} for an overview over existing methods for AF classification that are not based on deep neural networks. 
 
%\vspace{-0.15cm}
\section[Methods]{Methods\footnote{ Source code is available at:\\ {\tt https://github.com/yruffiner/ecg-classification}}}
\vspace{-0.15cm}
In this section we give a detailed description of our network architectures as well as the training and evaluation procedures used.
\vspace{-0.15cm}
\subsection{Network architectures} \vspace{-0.15cm}
We propose two neural network architectures for ECG classification, a CNN and a CRNN, illustrated in Fig. \ref{fig:architectures}. Both architectures consist of four parts: 1) data preprocessing computing a logarithmic spectrogram of the input; 2) a stack of convolutional layers for feature extraction; 3) aggregation of features across time by averaging and an LSTM block  in case of the CNN and the CRNN, respectively; 4) a linear classifier. In the following we describe each of the aforementioned parts in detail.

\begin{figure}[t]
\includegraphics[width=0.45\textwidth]{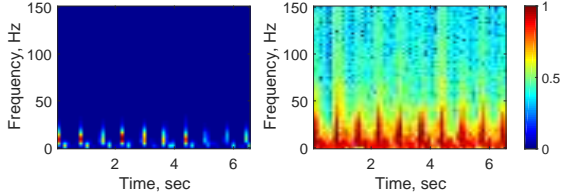}
\vspace{-0.1cm}
\caption{\label{fig:spectrogram}Normalized spectrogram (left) and normalized logarithmic spectrogram (right) of an example ECG signal.}
\vspace{-0.4cm}
\end{figure}

\bpar{1) Logarithmic spectrogram:} To preprocess the data we compute the one-sided spectrogram of the time-domain input ECG signal and apply a logarithmic transform. Preliminary experiments showed that the logarithmic transform considerably increases the classification accuracy; Fig.~\ref{fig:spectrogram} illustrates the effect of the logarithmic transform. The spectrogram is computed using a Tukey window of length 64 (corresponding to $213$ms at the $300$Hz sampling rate of the challenge data and resulting in $33$ effective frequency bins) with shape parameter $0.25$ and $50$\% overlap. 

\bpar{2) Convolutional layers:} All convolutional layers first apply a set of $5 \times 5$ convolutional filters, followed by Batch-normalization and ReLU activation. The convolutional layers are grouped in blocks of $4$ and $6$ layers for the CNN and CRNN architecture, respectively, referred to as ConvBlock4 and ConvBlock6. The number of channels (feature maps) as well as the size of the feature maps remains constant in all but the last layer of each ConvBlock. The last layer applies max-pooling over $2 \times 2$ windows and increases the number of channels. Specifically, the number of channels at the output of the first ConvBlock is $64$ and is increased by $32$ by each subsequent ConvBlock, resulting in $224$ $1$-dimensional and $160$ $3$-dimensional feature maps (per output time step) for the CNN and CRNN, respectively, at the output of the last ConvBlock fed to the feature aggregation part 3) (see Fig. \ref{fig:architectures}).

\bpar{3) Feature aggregation across time:}
As the ConvBlocks process the variable-length input ECG signals in full length, they produce variable length outputs, which have to be aggregated across time before they can be fed to a standard classifier (which typically requires the dimension of the input to be fixed). In our CNN architecture, temporal aggregation is achieved simply by averaging, whereas in the CRNN architecture the $3$-dimensional feature maps are first flattened and then feed to a $3$-layer bidirectional LSTM network with $200$ neurons in each layer. The (temporally) last output of the LSTM network then serves as the aggregated feature vector. 

Averaging realizes temporal smoothing of features and may therefore not be suited to classify episodic phenomena occurring only during a short time span relative to the signal length, as in certain types AF. The LSTM network, on the other hand, aggregates the features in a highly non-linear manner across time and potentially preserves episodic phenomena better. 

\bpar{4) Linear classifier:} We employ a standard linear layer with SoftMax to compute the class probabilities.

\newcommand{\dimsize}{\footnotesize}

\begin{figure}[ht]

\begin{subfigure}[t]{0.45\columnwidth}
\vskip 0pt
\begin{tikzpicture}[
    auto,
    decision/.style = { diamond, draw=blue, thick, fill=blue!20,
                        text width=5em, text badly centered,
                        inner sep=1pt, rounded corners },
    block/.style    = { rectangle, draw=blue, thick, 
                        fill=blue!20, text width=6em, text centered,
                        rounded corners, minimum height=2em },
    block1/.style    = { rectangle, draw=yellow, thick, 
                        fill=yellow!20, text width=6em, text centered,
                        rounded corners, minimum height=2em },
    block2/.style    = { rectangle, draw=red, thick, 
                        fill=red!20, text width=6em, text centered,
                        rounded corners, minimum height=2em },
    block3/.style    = { rectangle, draw=green, thick, 
                        fill=green!20, text width=6em, text centered,
                        rounded corners, minimum height=2em },                        
    line/.style     = { draw, thick, ->, shorten >=2pt },
    ]
  % Define nodes in a matrix
  \matrix [column sep=3mm, row sep=3mm] {
                    & \node [text centered] (x) {\textsf{ECG signal}};            & \\
                    & \node (null11) {}; & \\
                    & \node [block1] (doa) {\textsf{Log-Spectrogram}};   & \\
                    & \node (null1) {}; & \\
                    & \node [block] (uiddes) {\textsf{ConvBlock4}}; & \\
                    & \node (null12) {}; & \\
                    & \node (null121) {}; & \\
                    & \node [block] (track) {\textsf{ConvBlock4}}; & \\
                    & \node (null2) {}; & \\
                    & \node (null21) {}; & \\
                    & \node [block2] (pesos)
                        {\textsf{Temporal average}};            & \\
                    & \node (null13) {}; & \\
                    & \node (null131) {}; & \\[0.21cm]
                    & \node [block3] (filtrado)
                        {\textsf{Linear}};          & \\
                    & \node (null14) {}; & \\
                    & \node [text centered] (xf) {\textsf{Label} };          & \\
  };
  % connect all nodes defined above
  \begin{scope} [every path/.style=line]
    \path (x)        --    node [midway,anchor=base] {\colorbox{white}{\dimsize$[t\times 1 \times 1]$}} (doa);
    \path (doa)      --    node [midway,anchor=base] {\colorbox{white}{\dimsize$[t/32\times 33 \times 1]$}} (uiddes);
    \path[dashed] (uiddes)   --    node [midway,anchor=base] {\colorbox{white}{\dimsize$[t/64\times 17 \times 64]$}} (track);
    \path (track)    --    node [midway,anchor=base] {\colorbox{white}{\dimsize$[t/2048\times 1 \times 224]$}} (pesos);
    \path (pesos)    --    node [midway,anchor=base] {\colorbox{white}{\dimsize$[1\times 1 \times 224]$}} (filtrado);
    \path (filtrado) --    node [midway,anchor=base] {\colorbox{white}{\dimsize$[1\times 1 \times 4]$}}(xf);
    \draw [decorate,decoration={brace,amplitude=15pt,mirror,raise=34pt},yshift=0pt]
    (null1) -- (null2) node [black,midway,xshift=-2.4cm] {$\times 6$};
  \end{scope}
  
\end{tikzpicture}
\end{subfigure}
\hspace{0.5cm}
\begin{subfigure}[t]{0.45\columnwidth}
\vskip 0pt
\begin{tikzpicture}[
    auto,
    decision/.style = { diamond, draw=blue, thick, fill=blue!20,
                        text width=5em, text badly centered,
                        inner sep=1pt, rounded corners },
    block/.style    = { rectangle, draw=blue, thick, 
                        fill=blue!20, text width=6em, text centered,
                        rounded corners, minimum height=2em },
    block1/.style    = { rectangle, draw=yellow, thick, 
                        fill=yellow!20, text width=6em, text centered,
                        rounded corners, minimum height=2em },
    block2/.style    = { rectangle, draw=red, thick, 
                        fill=red!20, text width=6em, text centered,
                        rounded corners, minimum height=2em },
    block3/.style    = { rectangle, draw=green, thick, 
                        fill=green!20, text width=6em, text centered,
                        rounded corners, minimum height=2em },                        
    line/.style     = { draw, thick, ->, shorten >=2pt },
    ]
  % Define nodes in a matrix
  \matrix [column sep=3mm, row sep=3mm] {
                    & \node [text centered] (x) {\textsf{ECG signal}};            & \\
                    & \node (null11) {}; & \\
                    & \node [block1] (doa) {\textsf{Log-Spectrogram}};   & \\
                    & \node (null1) {}; & \\
                    & \node [block] (uiddes) {\textsf{ConvBlock6}}; & \\
                    & \node (null12) {}; & \\
                    & \node (null121) {}; & \\
                    & \node [block] (track) {\textsf{ConvBlock6}}; & \\
                    & \node (null2) {}; & \\
                    & \node [block2] (flatten)
                        {\textsf{Flatten}};          & \\
                    & \node (null13) {}; & \\
                    & \node [block2] (pesos)
                        {\textsf{LSTM Block}};            & \\
                    & \node (null13) {}; & \\
                    & \node [block3] (filtrado)
                        {\textsf{Linear}};          & \\
                    & \node (null14) {}; & \\
                    & \node [text centered] (xf) {\textsf{Label} };          & \\
  };
  % connect all nodes defined above
  \begin{scope} [every path/.style=line]
    \path (x)        --    node [midway,anchor=base] {\colorbox{white}{\dimsize$[t\times 1 \times 1]$}} (doa);
    \path (doa)      --    node [midway,anchor=base] {\colorbox{white}{\dimsize$[t/32\times 33 \times 1]$}} (uiddes);
    \path[dashed] (uiddes)   --    node [midway,anchor=base] {\colorbox{white}{\dimsize$[t/64\times 17 \times 64]$}} (track);
    \path (track)    --    node [midway,anchor=base] {\colorbox{white}{\dimsize$[t/512\times 3 \times 160]$}} (flatten);
    \path (flatten)    --    node [midway,anchor=base] {\colorbox{white}{\dimsize$[t/512\times 1 \times 480]$}} (pesos);
    \path (pesos)    --    node [midway,anchor=base] {\colorbox{white}{\dimsize$[1\times 1 \times 200]$}} (filtrado);
    \path (filtrado) --    node [midway,anchor=base] {\colorbox{white}{\dimsize$[1\times 1 \times 4]$}}(xf);
    \draw [decorate,decoration={brace,amplitude=15pt,mirror,raise=34pt},yshift=0pt]
    (null1) -- (null2) node [black,midway,xshift=-2.4cm] {$\times 4$};
  \end{scope}
  
\end{tikzpicture}
\end{subfigure}
\caption{The proposed CNN (left) and CRNN (right) architecture. The tensor dimensions are given in the format $[\text{time}\; t \times \text{nbr. of features} \times \text{nbr. of channels}]$. }
\label{fig:architectures}
\end{figure}
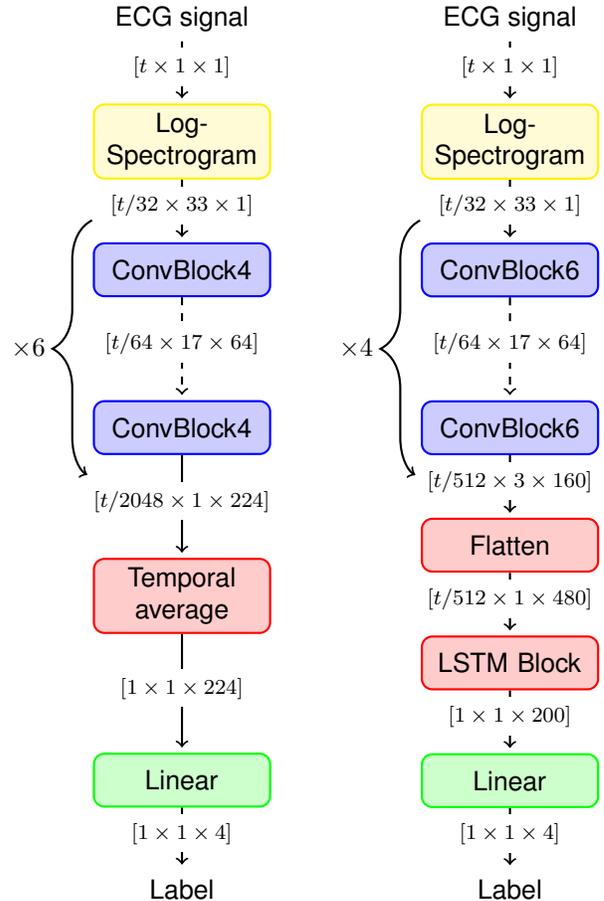

\subsection{Training} \label{sec:training} 

For both network architectures we used the cross-entropy loss (reweighted as to account for the class frequencies) as training objective, and employed the Adam optimizer with the default parameters recommended in \cite{kingmaB14}. The batch size was set to $20$. Furthermore, we used dropout with probability $0.15$ in all layers and early stopping based on the $F_1$ measure described in Sec. \ref{sec:evaluation}.

\bpar{Training protocols:} We trained the CNN end-to-end from scratch without encountering any issues. Training the convolutional and recurrent layers in the CRNN jointly from scratch, on the other hand, did not lead to convergence. We therefore adopted the following $3$-phase protocol to train the CRNN. In phase $1$, the LSTM block was replaced by feature averaging across time and the convolutional layers were trained together with a linear classifier for $500$ epochs. In phase $2$, the feature averaging operator was swapped with the LSTM block and the recurrent layers were trained for $100$ epochs, while keeping the convolution layers fixed. In phase $3$, the convolutional and recurrent layers were trained jointly, reducing the learning rate by a factor of $10$ every $200$ epochs.

\bpar{Data augmentation:} We observed severe overfitting in preliminary experiments. This can be attributed to the fact that number of parameters in the proposed architectures is large compared to the size of data set used for evaluation (see Sec. \ref{sec:evaluation}). It was demonstrated in \cite{simard2013augmentation} that data augmentation can act as a regularizer to prevent overfitting in neural networks, and also improves classification performance in problems with imbalanced class frequencies \cite{chawla2002augmentation}. We therefore developed a simple data augmentation scheme tailored to the ECG data at hand. Specifically, we employ two data augmentation techniques, namely \textit{dropout bursts} and \textit{random resampling}. 

Dropout bursts are created by selecting time instants uniformly at random and setting the ECG signal values in a $50$ms vicinity of those time instants to $0$. Dropout burst hence model short periods of weak signal due to, e.g., bad contact of ECG leads. 

Assuming a heart rate of $80$bpm for all training ECG signals, random resampling emulates a broader range of heart rates by uniformly resampling the ECG signals such that the heart rate of the resampled signal is uniformly distributed on the interval $[60, 120]$bpm. These emulated heart rates may be unrealistically high or low due to the assumption of an $80$bpm heart rate independently of the signal.

\bpar{Ensembling:} To exploit the entire data set at hand (recall that we employ early stopping which uses part of the data set for validation) we used ensembles of $5$ networks of the same type (i.e., either CNN or CRNN) to build production models, combining the individual predictions by majority voting. Specifically, we partitioned---in a stratified manner---the data set into $5$ equally sized subsets, and, for every network in the ensemble, used $4$ of the subset for training and the remaining subset for validation, choosing a different subset for validation for every network.

\subsection{Evaluation} \label{sec:evaluation}
We evaluated the proposed CNN and CRNN architecture on the publicly available PhysioNet/CinC Challenge 2017 data set containing 8,528 single lead ECG recordings of length ranging from $9$ to $61$sec, sampled at 300Hz. Each recording is labeled with one of the classes ``normal rhythm'', ``AF rhythm'', ``other rhythm'', and ``noisy recording'' (we will henceforth use the abbreviations ``N'', ``A'', ``O'', and ``\textasciitilde'', respectively). The classification performance was measured using the average over the class $F_1$ scores of the classes N, A, and O, i.e., $F_{1, \text{avg}} = \frac{1}{3} \sum_{c \in \{\text{N, A, O}\}} F_{1, c}$, where $F_{1, c} = 2 \text{\#TP}_c/(2 \text{\#TP}_c + \text{\#FN}_c + \text{\#FP}_c)$ (using $\text{TP}_c$, $\text{FP}_c$, and $\text{FN}_c$ to denote the true positives, false positives, and false negatives, respectively, for class $c$). 
We refer the reader to \cite{clifford2017af} for a detailed description of the data set. We evaluated the proposed network architectures via stratified $5$-fold cross-validation. To realize early stopping, for every fold, we split the training data into two partitions, one for training and one for validation containing $5/6$ and $1/6$, respectively, of the training data. Thus, for every fold, the effective training set size amounted to $4/5 \cdot 5/6 = 2/3$ or 66.6\% of all data available. We hence expect that an ensemble of $5$ networks yields a higher $F_{1, \text{avg}}$ as it exploits all data available.

To demonstrate the effectiveness of the proposed data augmentation scheme, we trained the CNN and CRNN exactly as described in Sec. \ref{sec:training}, but without data augmentation.

\begin{table}[t]
	\centering
\begin{tabular}{l l c c c c c}
	Arch. & metric & N & A & O & \textasciitilde & overall \\
	\hline 
	\multirow{2}{*}{CNN} & acc. & 88.1 & 83.6 & 66.9 & 77.1 & 81.2  \\ \cline{2-7}
	& $F_1$ & 87.8 & 79.0 & 70.1 & 65.3 & 79.0  \\ \hline \hline
	\multirow{2}{*}{CRNN} & acc. & 89.9 & 77.8 & 69.4 & 71.5 & 82.3  \\ \cline{2-7}
	& $F_1$  & 88.8 & 76.4 & 72.6 & 64.5 & 79.2 \\ \hline
\end{tabular}
	\caption{\label{tbl:resultswaug}Accuracies (acc.) and $F_1$ scores (in \%) for the proposed network architectures (estimated using 5-fold cross validation).}
\vspace{0.4cm}
	\centering
\begin{tabular}{l l c c c c c}
	Arch. & metric & N & A & O & \textasciitilde & overall \\
	\hline
	\multirow{2}{*}{CNN} & acc. & 90.5 & 64.2 & 68.0 & 54.9 & 80.5  \\ \cline{2-7}
	& $F_1$ & 88.3 & 69.9 & 69.1 & 59.6 & 75.8  \\ \hline \hline
	\multirow{2}{*}{CRNN} & acc. & 90.2 & 69.1 & 63.0 & 51.1 & 79.2  \\ \cline{2-7}
	& $F_1$ & 87.4 & 69.9 & 66.5 & 54.9 & 74.6 \\ \hline
\end{tabular}
	\caption{\label{tbl:resultsnoaug}Accuracies (acc.) and $F_1$ scores (in \%) for the proposed network architectures with \textit{data augmentation deactivated} (estimated using 5-fold cross validation).}
\end{table}

\section{Results} \label{sec:results}

Tables \ref{tbl:resultswaug} and \ref{tbl:resultsnoaug} show the class $F_1$ scores and $F_{1, \text{avg}}$ (overall) along with the corresponding classification accuracies for the proposed architectures with and without data augmentation, respectively. The CRNN yielded a higher overall accuracy and slightly higher $F_{1, \text{avg}}$ than the CNN when data augmentation was employed. The opposite can be observed in the case when data augmentation was deactivated. In both cases, none of the architectures has consistently higher class accuracies or class $F_1$ scores than the other. Data augmentation is seen to considerably increase $F_{1, \text{avg}}$ for both CNN and CRNN, with a slightly better improvement for the CRNN. 

Based on these results we chose to submit an ensemble of CRNNs to the PhysioNet/CinC Challenge 2017. This ensemble obtained an $F_{1, \text{avg}}$ of 0.82 on the private challenge testing set, which corresponds to the second best score (after rounding to two decimal places as per \cite{clifford2017af}) obtained in the challenge. In terms of running time, the ensemble on average consumed $58.1$\% of the computation quota available on the challenge evaluation server.

\section{Discussion}

The results presented in Sec.~\ref{sec:results} indicate that aggregation of features across time using an LSTM network is more effective than averaging in the ECG classification task under consideration, when data augmentation is employed. However, this has to be taken with a grain of salt as the CRNN has more parameters, and thereby potentially a higher model capacity, than the CNN. In addition, we observed that phase 3 of the CRNN training protocol did not consistently lead to an increase in $F_{1, \text{avg}}$, and further improvements might be achieved by refining the training protocol. Furthermore, the results in Sec.~\ref{sec:results} also show the effectiveness of the proposed data augmentation scheme, indicating that it captures certain real world phenomena---at least to some extent. 

We briefly comment on directions we explored in preliminary experiments, but which did not lead to  improvements and were therefore not included in our final training protocols. As an alternative to data augmentation we tried to pretrain the CNN and the convolutional layers of the CRNN on the PTB Diagnostic ECG Database \cite{bousseljot1995nutzung}, which contains $549$ $14$-lead ECG recordings of $290$ subjects with a variety of different cardiac conditions. This pretraining procedure did not lead to improvements compared to initialization with random weights. 
We further explored \cite[Alg. 7.3]{goodfellow2016deep} to incorporate the knowledge in the validation set into a single production model, which is more effective than ensembling from a computational and storage point of view. In a nutshell, \cite[Alg. 7.3]{goodfellow2016deep} continues training on the union of the training and the validation set after activation of early stopping until the average loss on the validation set attains the average loss on the training set obtained at the time of activation of early stopping. However, continuing training according \cite[Alg. 7.3]{goodfellow2016deep} led to a decrease in $F_{1, \text{avg}}$ in our challenge submissions.

\section{Conclusion}

We developed and evaluated two deep neural network architectures for ECG classification. In addition, we proposed a simple data augmentation scheme for ECG data and demonstrated its effectiveness. Applying our architectures to multi lead ECG data, possibly with different pathology, as well as refining and extending the data augmentation scheme, e.g., by taking the actual heart rate into account for random resampling (instead of assuming $80$bpm), are interesting directions to be explored in the future.

\vspace{0.2cm}

% \section*{Acknowledgements}  
% This section is not numbered.
% 
\noindent\bpar{Acknowledgments.}  
The Titan X Pascal GPU used for this research was donated by the NVIDIA Corporation.

% LateX generates the ``References'' heading automatically and switches
% to 9 point type for the bibliography.  If you use BibTeX (recommended),
% follow the examples in the sample 'refs.bib' file to enter your references,
% and leave the following line unchanged.
\bibliography{refs}
% LaTeX inserts the ``Address for correspondence'' heading.
\vspace{-0.1cm}
\begin{correspondence}
Dmytro Perekrestenko, Michael Tschannen\\
ETH Z{\"u}rich, Communication Technology Laboratory\\
Sternwartstrasse 7\\
CH-8092 Z{\"u}rich\\
Switzerland\\
\{pdmytro, michaelt\}@nari.ee.ethz.ch
\end{correspondence}

\end{document}